\documentclass[11pt, letterpaper]{article}
\usepackage[utf8]{inputenc}
\usepackage[margin=1.5cm]{geometry}
\usepackage{titlesec}
\usepackage{tabu}
\usepackage{enumitem}
\usepackage{amssymb}
\usepackage{xcolor}
\newlist{selectlist}{itemize}{2}
\setlist[selectlist]{label=$\square$,leftmargin=*,noitemsep,topsep=0pt}
\usepackage{comment}
\usepackage{amsmath,graphicx}
\usepackage{amsfonts}
\usepackage{lmodern}
\usepackage{etoolbox}
\patchcmd{\thebibliography}{\section*{\refname}}{}{}{}
\usepackage{algorithm}
\usepackage[noend]{algpseudocode}
\usepackage{hyperref}
\hypersetup{
    colorlinks=true,
    linkcolor=blue,
    filecolor=magenta,      
    urlcolor=blue,
}

\newcommand{\revision}[1]{\textcolor{blue}}

\titleformat{\section}[block]{\hspace{1em}\bfseries}{\thesection.}{0.5em}{} 
\titleformat{\subsection}[block]{\hspace{1em}}{\thesubsection}{0.5em}{}

\begin{document}

\noindent
\textbf{\textit{INCLG: Inpainting for Non-Cleft Lip Generation with a Multi-Task Image Processing Network}}

\vskip0.5cm
\noindent
\textbf{\textit{Shuang Chen (Durham University, shuang.chen@durham.ac.uk), \\
Amir Atapour-Abarghouei (Durham University, amir.atapour-abarghouei@durham.ac.uk), \\
Edmond S. L. Ho (University of Glasgow, shu-lim.ho@glasgow.ac.uk), \\
Hubert P. H. Shum (Durham University, hubert.shum@durham.ac.uk, corresponding author)}}\\

\noindent
\textbf{Abstract}\\
\textit{
We present a software that predicts non-cleft facial images for patients with cleft lip, thereby facilitating the understanding, awareness and discussion of cleft lip surgeries. To protect patients’ privacy, we design a software framework using image inpainting, which does not require cleft lip images for training, thereby mitigating the risk of model leakage. We implement a novel multi-task architecture that predicts both the non-cleft facial image and facial landmarks, resulting in better performance as evaluated by surgeons. The software is implemented with PyTorch and is usable with consumer-level color images with a fast prediction speed, enabling effective deployment.
}
\vskip0.5cm

\noindent
\textbf{Keywords}\\
\textit{Cleft Lip; Image Inpainting; Deep Neural Network; Multi-task Learning; Face Modelling}
\vskip0.5cm
\newpage
\noindent
\textbf{Code metadata}\\

\noindent
\begin{tabular}{|l|p{6.5cm}|p{9.5cm}|}
\hline
\textbf{Nr.} & \textbf{Code metadata description} & \textbf{Please fill in this column} \\
\hline
C1 & Current code version & v1 \\
\hline
C2 & Permanent link to code/repository used for this code version & 
\href{https://github.com/ChrisChen1023/INCLG}{https://github.com/ChrisChen1023/INCLG} \\
\hline
C3  & Permanent link to Reproducible Capsule & \href{https://codeocean.com/capsule/4388343/tree/v1}{https://codeocean.com/capsule/4388343/tree/v1} \\
\hline
C4 & Legal Code License   & MIT License \\
\hline
C5 & Code versioning system used & git \\
\hline
C6 & Software code languages, tools, and services used & Python \\
\hline
C7 & Compilation requirements, operating environments \& dependencies & Python 3.8, PyTorch 1.10.2, torchvision 0.11.3\\
\hline
C8 & If available Link to developer documentation/manual & \href{https://codeocean.com/capsule/4388343/tree/v1}{https://codeocean.com/capsule/4388343/tree/v1} \\
\hline
C9 & Support email for questions & shuang.chen@durham.ac.uk \\
\hline
\end{tabular}\\

\vskip0.5cm
\noindent
\section{Introduction}

A cleft lip/palate is a medical condition where the lip/palate of a patient does not join completely before birth, which usually occurs in the early stages of pregnancy. In the UK, cleft lips are the most common facial birth defect, with one out of every 700 children suffering from cleft lip and palate every year ~\cite{cleft_lip_NHS}. This explains the importance of cleft lip and palate surgeries, which are usually performed on orofacial cleft patients at an average age of three months \cite{wellens2006keys}. Although the surgical treatment for cleft lip and palate varies, their common objective is to achieve symmetry and enhance a nasolabial look~\cite{mosmuller2017development}. 

As cleft lips are pre-born defects, many parents of the patients would find it hard to imagine what the non-cleft faces of their children would be like. To facilitate the understanding, awareness and discussion of cleft lip surgeries, we have worked with the UK's Royal Victoria Infirmary (RVI) to collect a dataset of cleft lips patients, and designed a system that allows the prediction of non-cleft faces from the cleft lip counterparts. 


A core challenge of our software design is to protect the privacy of cleft lip patients. Research has shown that due to the high memory capacity of deep learning models, it is possible to reconstruct original training samples from the network parameters, a scenario known as model leakage \cite{zhu2019deep}. While it may be straightforward to formulate the non-cleft facial-prediction task using a style transfer \cite{8237506} framework, training such systems requires both cleft and non-cleft facial images, resulting in a risk of model leakage. We present a novel software engineering design by tackling non-cleft facial-prediction with an image in-painting framework. This allows us to train the system using open facial datasets \cite{liu2015deep} with a tailor-made algorithm to mask out the mouth area, and test the system with cleft lip images. As a result, the model parameters do not store any cleft lip information.

In particular, we develop a PyTorch-based software that utilizes a state-of-the-art image inpainting network \cite{yang2019lafin} as the backbone, and develop a multi-task system that predicts both images and the facial landmarks, thereby generating non-cleft faces. The facial landmark task provides geometric information that facilitates the image generation task. 
Compared to existing work that utilizes a multi-stage framework to first predict landmarks and then predict images \cite{Nazeri_2019_ICCV, yang2019lafin}, ours is superior as both tasks are performed at the same time, avoiding any error propagation from the first stage to the second stage. 

The quality of images produced by our software has been evaluated by NHS surgeons, showcasing its superior performance to alternative designs. It has a fast inference speed and works with color images captured by consumer-level cameras, allowing an effective deployment process. It is open-source, facilitating research and development in this area.

The source code presented in this paper has been originally developed to implement the theory proposed in \cite{chen2022feasibility}, which is accepted in a biomedicine-focused conference. 
In this paper, we explain the implementation details of this software and its impact in the real world. In particular, we focus on the design concepts of the software architecture and the details of the engineering considerations. This is further supported by a validated version of the source code in the CodeOcean environment.

\section{System description}
To protect the privacy of patients' data, we decide to implement the non-cleft facial image prediction system as an image inpainting framework. One key software engineering decision in this research is the framework we use to implement the solution. Existing style transfer-based frameworks \cite{8237506} allow effective facial image generation with different features. However, they require training data from both the source (i.e. cleft lip images in our case) and target (i.e. non-cleft lip images) domains, which may lead to model leakage where the trained model memorizes the training images. Conditional image translation frameworks using GAN \cite{8237506} or VAEs \cite{nozawa21car} may resolve the issue, but those methods mainly focus on the synthesis of new color patterns instead of geometric structures. Our investigation led us to the image inpainting framework \cite{yang2019lafin} as a suitable solution, as it does not necessitate using cleft facial data for training. Additionally, the binary mask effectively defines the lip area for synthesis with the rest of the face, serving as conditions, making it well-suited to our requirements.

In particular, to implement an image inpainting framework, we utilize the image generation network in ~\cite{yang2019lafin} as the backbone, which is ameliorated from ~\cite{Nazeri_2019_ICCV}, given its good performance in image inpainting. We also re-implemented the gated convolution algorithm proposed in~\cite{yu2019free} to dynamically select features for each channel and location, resulting in better inpainting quality. 

On top of the backbone, we implement a multi-task system that predicts both the non-cleft facial image and facial landmarks. Facial landmark has shown to be effective in assisting facial image inpainting \cite{Nazeri_2019_ICCV, yang2019lafin}, and is used extensively for cleft lip analysis \cite{li2019clpnet}. Our work differs from existing approaches in that we employ a multi-task model, where two tasks share a part of a common network and facilitate each other.

To prepare the training data, we employ an open facial dataset and a tailor-made masking algorithm. In particular, we use the CelebA dataset \cite{liu2015deep}, which consists of 202,599 face images of over 10,000 celebrities. To prepare the data for training our inpainting network, we apply an irregular mask algorithm following \cite{liu2018image}, such that our network can learn to inpaint any masked regions of the face.

To test the system, we work with the NHS to collect a dataset of cleft lip images. Due to the sensitive nature of the data, ethical approvals are obtained from the Research Ethics Committee (REC), the Health Research Authority (HRA), and Health and Care Research Wales (HCRW), under Approval Nos. 19/LO/1690 and under IRAS Project ID: 240451. Given a cleft lip image, we manually draw a mask that covers the mouth area. The masked image is fed into our multi-task network to create the non-cleft facial counterpart, with the facial landmark as a side-product. Since cleft lip images are only used in testing, we mitigate any risk of model leakage \cite{zhu2019deep}.


\subsection{Network Design and Implementation}
\begin{figure}
\includegraphics[width=\textwidth]{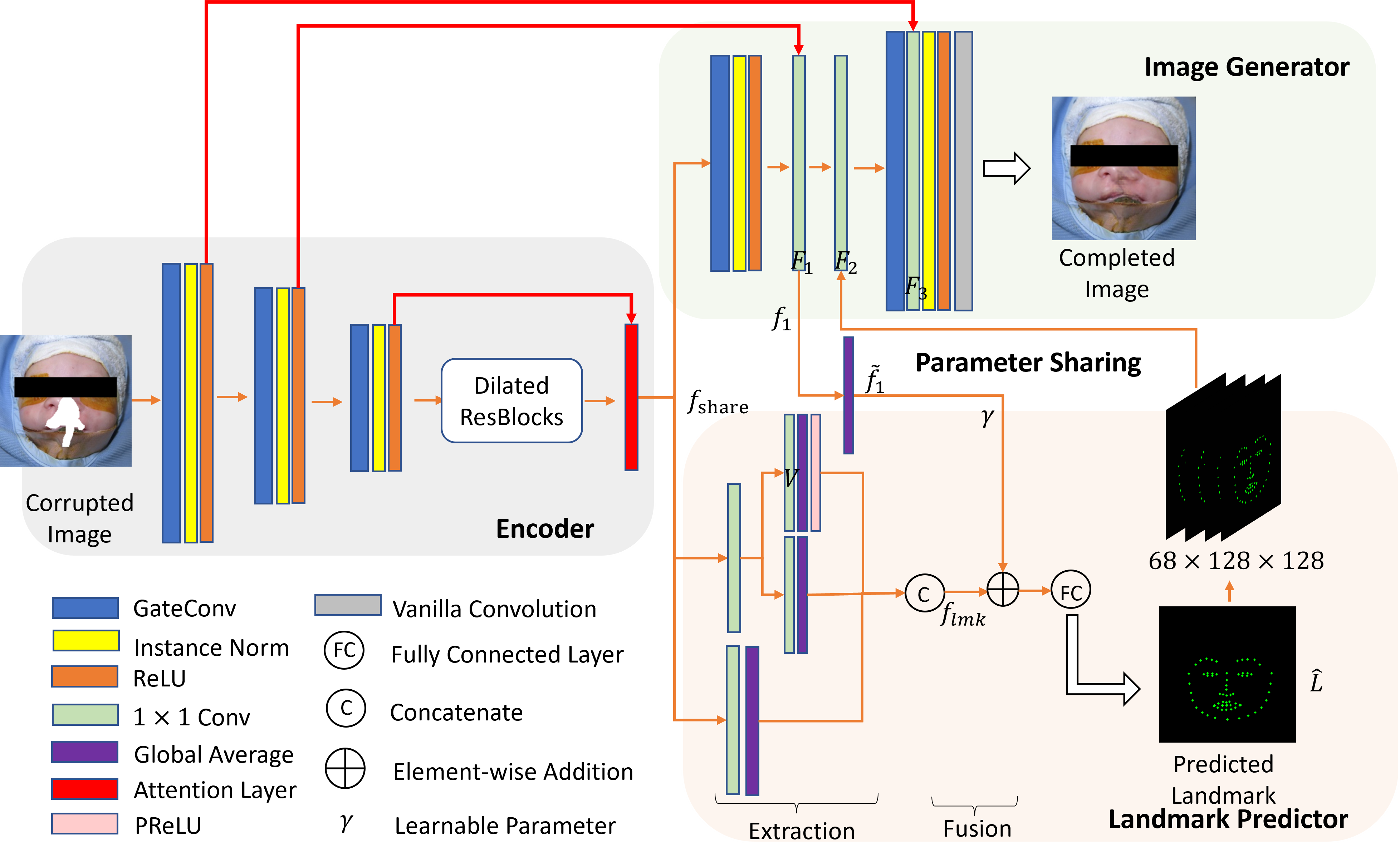}
\caption{The overview of proposed multi-task architecture.}
\label{fig1}
\end{figure}

Here, we provide details for the design and implementation of our deep neural network, as shown in Figure \ref{fig1}.

The encoder is used to encode an image into a feature representation. We develop a gated convolution block that includes a gated convolution layer \cite{yu2019free}, a normalisation layer and an activation layer (ReLU). A masked image is fed into three gated convolution blocks with decreasing feature sizes from $256\times256$ to $64\times64$. 
Subsequently, the encoded feature is passed into multiple dilated residual blocks to extend the receptive field of the encoder. 
At the end of the encoder, we follow ~\cite{zheng2019pluralistic} to implement an attention mechanism to match the masked and unmasked regions. 
After a skip connection, the encoder outputs the shared feature map $f_\text{share}$, which is practically a concise representation of the image. The feature map is passed to both the image generator and predictor.

The image generator is used to predict the non-cleft facial image. Given the shared feature $f_\text{share}$, we employ a gated convolution block to implement upsampling. This is followed by two $1\times1$ convolution layers, $F_1$ and $F_2$, for feature fusion. The first one is utilised to fuse the encoder feature with the skip connection, while the second one is responsible for parameter sharing to fuse the landmark indicator. This is followed by another upsampling gated convolution block with a fusion layer $F_3$ and a convolutional layer to synthesise the non-cleft lip image. 


The landmark predictor is used to predict the landmark of the non-cleft facial image. Following \cite{yang2019lafin}, the shared feature $f_\textbf{share}$ is passed into different $1\times1$ convolution layers followed by global average pooling to extract features of different numbers of channels. The feature with the largest number of channels (i.e. $V$) is further passed into a $\text{PReLU}$~\cite{he2015delving} activation layer. These features are concatenated and fused with the image features to predict the landmark.

We develop a parameter-sharing mechanism to share information between the image generator and the landmark predictor. We implement an adaptive feature fusion algorithm, in which the image feature $f_1$ from the layer $F_1$ is passed from the image generator to the landmark predictor. This is followed by a fully connected layer to generate the 68 landmark points, $\hat{L}$:  
\begin{equation}
\hat{L}=FC(\gamma * \widetilde{f}_{1} \oplus f_{l m k}),
\label{adaptive}
\end{equation}
where  $\gamma$ is a trainable parameter with zero initialization, $\oplus$ is element-wise addition, $\widetilde{f}_{1}$ is obtained by passing $f_1$ through a global average pooling layer and $f_{l m k}$ is the extracted landmark feature map. 
The predicted landmark points $\hat{L}$ are mapped into a $128\times128$ image corresponding to the landmark position. The image is stacked channel-wise to increase its influence (68 times in our setup), and passed from the landmark predictor back to $F_2$ in the image generator.

\begin{algorithm}
\caption{GAN-based training for proposed Multi-task model $MT_\theta$ {This is the new pseudo code}} 
\textbf{Inputs}: Generated image $x$, image ground truth $X$, predicted landmark $k$, landmark ground truth $K$, irregular mask $M$, max number of iterations $T$, batch size = $4$.\\
\textbf{Output}: Multi-task model $MT_\theta$

\begin{algorithmic}[1]
\State Build dataloader.
\State Initialize the network.
\If{t $<$ T}:
\State Sample $4$ images and corresponding landmark from dataloader.
\State Sample $4$ irregular mask from dataloader.
\State $x, k, L_\text{pixel}, L_\text{landmark}, L_\text{tv}, L_\text{style}, L_\text{perceptual}, L_g, L_d  \gets MT_\theta(X,M,L)$.
\State $L_G \gets L_\text{pixel} + L_\text{landmark} + L_\text{tv} + L_\text{style} + L_\text{perceptual} + L_g$ 
\State $L_D \gets L_d$
\State Freeze the $\theta_G$ in multi-task model, update discriminator with adversarial loss $L_D$.
\State Freeze the $\theta_D$ in discriminator, update multi-task model with adversarial loss $L_G$.
\State $t \gets t+1$
\EndIf
\State Save the trained Multi-task model $MT_\theta$.
\end{algorithmic}
\label{algorithm1}
\end{algorithm}

\subsection{Implementation details}
This multi-tasking image inpainting system is programmed in PyTorch. The main packages include numpy 1.15.4, torch 1.10.2 and torchvision 0.11.3. All of our experiments are implemented with $256\times256$ images and masks. We employ the $\text{imageio 2.15.0}$ to load and $\text{Pillow 8.4.0}$ to resize both images and masks. For each facial image, Face Align Network (FAN) generates $136$ values to denote the x- and y- positions of the 68 landmark points.

To train the network, a GAN-based~\cite{GAN2014} training flow is employed as shown in Algorithm~\ref{algorithm1}. We first apply FAN~\cite{bulat2017far}, a Face Align Network, to generate the ground truth landmark points from CelebA~\cite{liu2015deep}, following the default training and testing split of the dataset. 
We then apply Optuna~\cite{akiba2019optuna} for hyperparameter tuning. Specifically, we fix all hyperparameters except the weight of landmark loss, and train our model with one epoch using Optuna to sample such weights. 
With the same method, we also tune other hyperparameters, such as the learning rate, the decay weight of the learning rate and batch size. 

The collected dataset is used for inference. From both quantitative and qualitative results, our system generates semantically plausible non-cleft facial images~\cite{chen2022feasibility}. 
The results are further evaluated by cleft lip surgeons, showcasing that our proposed network generates better images than state-of-the-arts \cite{guo2021image, Nazeri_2019_ICCV, yang2019lafin}.

The run-time cost of the proposed system is very low. Using our real-world cleft lip data, the inference step is implemented using an NVIDIA GeForce GTX 970 on a laptop, with an inference time of 200ms for a single image. This means that our trained system does not require a particularly powerful computing system to perform the inference, and a standard workstation or laptop computer can use our system. For training the network, one NVIDIA TITAN Xp is used for four days, which is typical in deep learning applications of a similar scale. 

\subsection{{How to use}}

To retrieve the training dataset for this image inpainting application, users are required to download the CelebA Dataset~\cite{liu2015deep} and the irregular mask dataset~\cite{liu2018image} from the respective official websites. The CelebA dataset should then be divided into a standard training set and a validation set, according to the official instruction. Additionally, the corresponding landmark points should be generated with FAN~\cite{bulat2017far}. Furthermore, the irregular mask dataset should be divided into three groups according to the mask ratios (0-20\%, 20-40\%, 40-60\%). 3,300 masks are randomly selected from each group, resulting in a total of 9,900 mask images for training. Another 200 masks are selected from each group, resulting in a total of 600 mask images for verification. For the inference step, all cleft facial images and their corresponding masks serve as the image test set and the mask test set, respectively. The user should then run the provided “./scripts/filst.py” script to generate training, test and validation set file lists, and update the information in the “config.yml” file accordingly to set the model configuration. Once the python environment has been set up using the released “requirements.txt” file, the user may proceed to run the “train.py” script for training and the “test.py” script for testing. For the inference process, although we recommend using our system with GPUs for better speed, the system is fully runnable with only CPUs.
Due to the sensitivity of patient privacy, we are not allowed to upload the cleft lip data for an online demonstration. Therefore, we show the reproducibility of our system with the images from CelebA and the irregular masks.

\section{Impact overview}
While our method primarily focuses on cleft lips, the uses of the implemented source code can be extended to other applications. The key idea of this software is to mask out a particular region of a face, and to employ inpainting techniques for predicting the masked area. The versatility of our system allows for the implementation of extended facial applications, such as makeup and plastic surgery prediction. To utilize these capabilities, a customized dataset is required for training, such as the Facial Beauty Database~\cite{ZHANG2019339} or a plastic surgery facial dataset~\cite{9150598}. The users then need to retrain our model according to section 2.3. The resulting model can then be tested using a corresponding mask that covers specific facial components, such as nose or eyebrows, to generate the image of the subject after makeup or plastic surgery. Therefore, it can also be used for supporting plastic surgeries and makeup prediction \cite{organisciak20makeup} on specific facial components. This would facilitate the understanding and discussion of those operations and applications among stakeholders.


We put a particular effort in selecting a software framework that is robust against model leakage and attack \cite{tramer2016stealing,zhu2019deep}. In particular, we propose the idea of excluding patient data in training deep learning models if possible, mitigating any privacy concerns and risk of data loss. The high-level concept of training with open data and testing with sensitive data can be employed in other machine learning applications to protect data privacy, particularly those in the healthcare domain or involving people of vulnerable groups.

In theory, our system is also capable of synthesising cleft facial images from non-clelf lip ones. In practice, due to the wide variety of cleft lip conditions, training such a system would require a large dataset of cleft images, which is currently not available. Should there be enough data (and we only need the lip area to protect patients' privacy), this system can be used to generate synthetic cleft lip facial images, which enable the training of machine learning algorithms. As the data is artificially created, there is no privacy or model leakage concern, and an unlimited amount of samples can be created. This aligns with the recent trend of using computer graphics techniques to mock up real-world data \cite{Hesse:MICCAI:2018}, facilitating the training of machine learning systems for patient-related applications \cite{zhang22cerebral}. Since the beginning of this research, there is raising awareness from both UK universities and hospitals in collecting cleft lip data for research purposes. We believe our vision will be made possible in the future.


{\section{Conclusion}}
This work implements a multi-task image inpainting model to predict non-cleft lip facial images from cleft lip ones. We make an important software engineering decision to implement the system under an inpainting framework, which does not require patient data for training and mitigates model leakage risks. We design and develop a multi-task neural network that co-predicts a facial image and the corresponding facial landmarks, and we find that the two tasks support each other. Apart from detailing the design and implementation details of our software, we also discuss its impact within and beyond cleft lip applications. The source code is now publicly released on CodeOcean and Github.


\hspace*{\fill} 
\\
\noindent 
\textbf{Declaration of competing interest}\\
The authors declare that they have no known competing financial interests or personal relationships that could have appeared to influence the work reported in this paper.

\hspace*{\fill} 
\\
\noindent 
\hspace*{\fill} 
\\
\noindent 
\textbf{References}
\bibliographystyle{IEEEtran}
\bibliography{refs}

\end{document}